\documentclass[conference]{IEEEtran}
\IEEEoverridecommandlockouts
\usepackage[numbers]{natbib}
\usepackage{amsmath,amssymb,amsfonts}
\usepackage{algorithmic}
\usepackage{graphicx}
\usepackage{textcomp}
\usepackage{xcolor}
\usepackage{cleveref}
\usepackage[utf8]{inputenc}
\usepackage{soul}
\usepackage{tikz}
\newcommand{\tikzxmark}{%
\tikz[scale=0.23] {
    \draw[line width=0.7,line cap=round] (0,0) to [bend left=6] (1,1);
    \draw[line width=0.7,line cap=round] (0.2,0.95) to [bend right=3] (0.8,0.05);
}}
\newcommand{\tikzcmark}{%
\tikz[scale=0.23] {
    \draw[line width=0.7,line cap=round] (0.25,0) to [bend left=10] (1,1);
    \draw[line width=0.8,line cap=round] (0,0.35) to [bend right=1] (0.23,0);
}}
\usepackage[english, vietnamese]{babel}
\def\BibTeX{{\rm B\kern-.05em{\sc i\kern-.025em b}\kern-.08em
    T\kern-.1667em\lower.7ex\hbox{E}\kern-.125emX}}
    
\begin{document}

\title{Utilize Transformers for translating Wikipedia category names\\
}

\author{
\IEEEauthorblockN{1\textsuperscript{st} Hoang-Thang Ta}
\IEEEauthorblockA{
\textit{Department of Information Technology} \\
\textit{Dalat University} \\
Da Lat, Vietnam \\
thangth@dlu.edu.vn}
\and
\IEEEauthorblockN{2\textsuperscript{nd} Quoc Thang La}
\IEEEauthorblockA{
\textit{Department of Information Technology} \\
\textit{Dalat University} \\
Da Lat, Vietnam \\
thanglq@dlu.edu.vn}
}

\maketitle
\begin{otherlanguage}{english}

\begin{abstract}
On Wikipedia, articles are categorized to aid readers in navigating content efficiently. The manual creation of new categories can be laborious and time-intensive. To tackle this issue, we built language models to translate Wikipedia categories from English to Vietnamese with a dataset containing 15,000 English-Vietnamese category pairs. Subsequently, small to medium-scale Transformer pre-trained models with a sequence-to-sequence architecture were fine-tuned for category translation. The experiments revealed that OPUS-MT-en-vi surpassed other models, attaining the highest performance with a BLEU score of 0.73, despite its smaller model storage. We expect our paper to be an alternative solution for translation tasks with limited computer resources.
\end{abstract}


\section{Introduction}
As the largest online encyclopedia, Wikipedia relies on a vast community of editors to develop its daily content, including creating new categories. However, this task can be both tedious and repetitive. Applying neural language models offers a promising solution to streamline this process and alleviate the burden on editors. By automatically translating category names, these models can significantly reduce the time and effort required from editors. This not only enhances the efficiency of category creation but also ensures consistency and accuracy across different language versions of Wikipedia.

Nowadays, most machine translation systems use neural network models trained on billions of parameters and massive datasets to boost the outcome quality compared to other approaches, such as rule-based machine translation (RBMT) and statistical machine translation (EBMT). With large language models like ChatGPT~\cite{deng2022benefits} dominating the output quality in many natural language processing (NLP) tasks, we search for proper solutions for training models in an affordable deployment. Therefore, we fine-turn small-medium scale Transformer pre-trained models like BART-base, T5-small, and OPUS-MT-en-vi over our datasets, including 15000 English-Vietnamese category pairs split into training, validation, and test sets in an 8:1:1 ratio. The experiments show that OPUS-MT-en-vi achieves the best performance on the test set while having the smallest model storage. Our main contribution is building a proper model for effectively translating category names from English to Vietnamese in terms of computer resources and model storage.

Except for this section, we outline related works in \Cref{related_works} and introduce our collected dataset in \Cref{dataset}, then our methodology for training language models in \Cref{methodology}. We performed our experiments and gave comments on the results in \Cref{experiments} before making conclusions in \Cref{conclusion}.

\section{Literature Review}\label{related_works}
The current popular approach in machine translation involves utilizing modern neural networks, which are trained on extensive datasets containing millions to billions of parameters. This approach has proven to achieve substantial quality improvements. At the same time, traditional methods are now less commonly used due to their limitations in dealing with new domains and expensive cost and language pairs with significantly different word orders~\cite{okpor2014machine}.

Many works on neural machine translation rely on an encoder-decoder architecture~\cite{garg2018machine}. \citet{cho2014learning} introduced the RNN Encoder-Decoder with two RNN networks to improve phrase representation using conditional probabilities. This model captures semantically and syntactically meaningful representations of linguistic phrases. \citet{sutskever2014sequence} created a sequence-to-sequence network using multilayered LSTMs to encode input sequences into fixed-dimensional vectors and then decode them into target sequences. Their models effectively handle long sentences and capture coherent, word-order-sensitive representations. To overcome the limitation performance of using a fixed-length vector, \citet{bahdanau2014neural} extended the encoder-decoder model with an attention mechanism to automatically (soft-)search for relevant parts of a source sentence when predicting a target word. 

As a very popular model in machine translation, \citet{vaswani2017attention} introduced the Transformer with decoder and encoder units, which rely entirely on attention mechanisms, eliminating recurrence and convolutions. The encoder comprises six identical layers with a multi-head self-attention mechanism and a position-wise feed-forward network. The decoder has six identical layers, adding a third sub-layer for multi-head attention over the encoder's output. \citet{liu2020very} enhanced the performance of the Transformer model by constructing a network comprising 60 encoder layers and 12 decoder layers, achieving a state-of-the-art BLEU score of 46.4 on the WMT14 English-French. 

Wikipedia categories have become a research topic in many works.  \citet{nastase2008decoding} decoded Wikipedia category names to induce relations between concepts. This structure allows the propagation of detected relations to numerous concept links, supporting the idea that Wikipedia category names are a rich source of accurate knowledge. \citet{chernov2006extracting} suggest extracting semantic information from Wikipedia by analyzing category links. This can build a semantic schema to improve search capabilities and offer editing suggestions. Their analysis shows that the Connectivity Ratio correlates positively with semantic connection strength. For translating category names from English to Vietnamese, \citet{ta2017phan} described the category name structures in both languages (English and Vietnamese) and provided translated examples based on these structures. However, their rule-based approach is inflexible with diverse category structures. To address this limitation, we consider some Transformers that support Vietnamese letters, such as OPUS-MT-En-Vi~\cite{tiedemann2020opus}, MTet~\cite{ngo2022mtet} with two widely-use models (BART-base~\cite{lewis2019bart} and T5-base~\cite{raffel2020exploring}), for the translation.

\section{Dataset}\label{dataset}
We built a crawler to collect random English-Vietnamese category pairs from Wikidata. Initially, the crawler randomly generates a set of Wikidata item indexes, which each has the format \texttt{Q-xx} with the prefix \texttt{Q-} followed by an integer \texttt{xx}. Utilizing Wikidata APIs\footnote{https://www.wikidata.org/w/api.php}, it accesses the item and extracts category names in English and Vietnamese to create source-target pairs. Because some pre-trained models like T5-base or BART-base do not support Vietnamese letters, we created a simple function to convert 134 diacritic letters to the corresponding encoded letters starting with the prefix \texttt{@}s and their indexes, as shown in \Cref{tab_diacritic_letters}.

\begin{table}[htbp]
\caption{Some diacritic letters and their encoded ones.}
\begin{center}
\begin{tabular}{|c|c|c|}
\hline  
\textbf{Index} & \textbf{Letters} & \textbf{Encode letters} \\
\hline  
1 & À & @1 \\
2 & Á & @2 \\
3 & Â & @3 \\
... & ... & ... \\
133 & Ỹ & @133 \\
134 & ỹ & @134 \\
\hline
\end{tabular}
\label{tab_diacritic_letters}
\end{center}
\end{table}

The data collection process is repeated in parallel until 15000 pairs are gathered. Then, we randomly divided the collected dataset into train (12000 pairs), validation (1500 pairs), and test sets (1500 pairs) with a ratio of 8:1:1 to ensure the natural data distribution in all sets.  Next, we perform basic analyses on the whole dataset, shown in \Cref{tab_basic_analysis}. The maximum lengths for source and target sequences are 11 and 14, respectively. Therefore, we set the maximum length to 16 for both inputs and outputs during data training. The vocabulary sizes for the sources (English) and targets (Vietnamese) in both sensitive and insensitive cases range from 7000 to 8000 words, reflecting limited diversity in our dataset. Additionally, there are lists of 10 common and rare words for sources and targets. The rare words, in particular, highlight the challenge of the translation task. Our intuition suggests that if these rare words do not appear in the training set, they may lead to translation errors in the test set.

\begin{table}[htbp]
\caption{Some basic analyses over the whole dataset.}
\begin{center}
\begin{tabular}{|p{3cm}|p{4cm}|}
\hline  
\textbf{Feature} & \textbf{Value} \\
\hline  
Maximum length of sources  & 11 \\
\hline
Maximum length of targets & 14 \\
\hline
Vocabulary size in sources (sensitive) & 8587 \\
\hline
Vocabulary size in targets (sensitive) &  8035 \\
\hline
Vocabulary size in sources (insensitive) & 7890 \\
\hline
Vocabulary size in targets (insensitive) & 7362 \\
\hline
10 popular words in sources & establishments, people, country, deaths, century, united, history, births, states, templates \\
\hline
10 popular words in targets & \begin{otherlanguage}{vietnamese}khởi đầu, bóng đá, quốc gia, thập niên, thế kỷ, sinh, châu, thể thao, lịch sử, bản mẫu\end{otherlanguage} \\
\hline
10 rare words in sources & achoerodus, alone, weird, yankovic, paula, abdul, waiting, anonymity, areca, nomascus \\
\hline
10 rare words in targets & \begin{otherlanguage}{vietnamese}một mình, hình vuông, ma thuật, ẩn danh, vượn mào, chụp, màn hình, huyền thoại, sơ khai báo viết, achoerodus\end{otherlanguage} \\
\hline
\end{tabular}
\label{tab_basic_analysis}
\end{center}
\end{table}

\Cref{tab_examples} shows several examples in our training set; each contains English, Vietnamese, and Encoded Vietnamese category names. These texts are short, and there is a gap in word order between English and Vietnamese category names. 

\begin{table}[htbp]
\caption{Some examples in the training set.}
\begin{center}
\begin{tabular}{|p{2cm}|p{2cm}|p{2.5cm}|}
\hline  
\textbf{English (source)} & \textbf{Vietnamese (target)} & \textbf{Encoded Vietnamese (encoded\_target)} \\
\hline  
Human development &  \begin{otherlanguage}{vietnamese}Phát triển con người\end{otherlanguage} & Ph@17t tri@79n con ng@43@105i \\
\hline
History of Oslo & \begin{otherlanguage}{vietnamese}Lịch sử Oslo\end{otherlanguage} & L@87ch s@121 Oslo \\
\hline  
History of literature by country & \begin{otherlanguage}{vietnamese}Lịch sử văn học theo quốc gia\end{otherlanguage} & L@87ch s@121 v@33n h@89c theo qu@93c gia \\
\hline  
2004 horror films & \begin{otherlanguage}{vietnamese}Phim kinh dị năm 2004\end{otherlanguage} & Phim kinh d@87 n@33m 2004 \\
\hline 
Disney articles by importance & \begin{otherlanguage}{vietnamese}Bài viết về Disney theo độ quan trọng\end{otherlanguage}  & B@16i vi@75t v@77 Disney theo @35@101 quan tr@89ng \\
\hline 
\end{tabular}
\label{tab_examples}
\end{center}
\end{table}

\section{Methodology}\label{methodology}

\begin{figure}[htbp]
\includegraphics[width=0.45\textwidth]{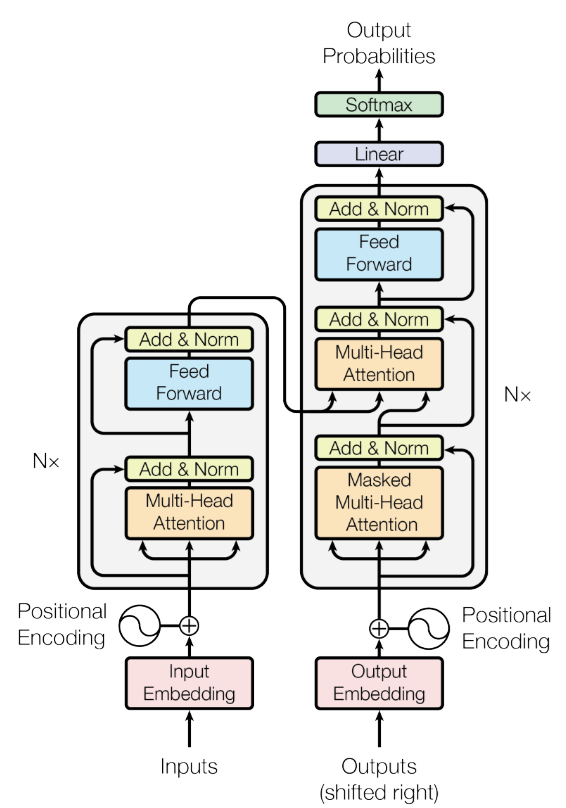}
\caption{The Transformer architecture~\cite{vaswani2017attention}.}
\label{fig_transformer}
\end{figure}

In this paper, we apply only Transformers, which follow a sequence-to-sequence architecture and contain an encoder, a decoder, and an attention mechanism. \Cref{fig_transformer} shows the typical architecture of a Transformer. Given a source $X=\{x_1, x_2, x_3, ..., x_N\}$ with $N$ symbols and a target $Y=\{y_1, y_2, y_3,...,y_M\}$ with $M$ symbols. The encoder yields a representation $Z = \{z_1, z_2, z_3,..., z_N\}$ from $X$ with the same number of symbols. Later, the decoder takes $Z$ to produce the target $Y$. The chain rule probability $p(Y|Z)$ to generate $Y$ from $Z$ is:
\begin{equation}
\begin{aligned}
p(Y|Z) &= \prod_{i}^{M}p(y_i|Y_{<i},Z)
\end{aligned}
\label{eq:chain_prob}
\end{equation}
which $y_0$ is the ``start'' symbol (\texttt{<bos>}) and $Y_{<i}$ is a sequence of previous symbols of $y_i$. When meeting the      ``end'' token (\texttt{<eos>}) or the maximum length, the inference process ends. The cross-entropy loss $L_{ent}$ minimizes the sum of negative loglikelihoods of the symbols:
\begin{equation}
\begin{aligned}
L_{ent} = - \sum^{M}_{j=1}\sum^{}_{w}p_{true}(w|Y_{<j},Z)log( p(w|Y_{<j},Z))
\end{aligned}
\label{eq:entropy_loss}
\end{equation}
which $p_{true}$ is a one-hot distribution:

\begin{equation}
\begin{aligned}
p_{true}(w|Y_{<j},Z) = \bigg\{
\begin{matrix}
1 \;\;\;\;\; w = y_j 
\\
0 \;\;\;\;\; w \neq y_j
\end{matrix}
\end{aligned}
\label{eq:one_hot_dis}
\end{equation}

A Transformer has two attention functions: Scaled Dot-Product Attention and Multi-Head Attention. Let $Q$, $K$, $V$ be the query matrix, the key matrix, and the value matrix correspondingly. Let $d_k$, $d_k$ be the dimensions of queries and keys, and $d_v$ be the dimension of values. The attention function of Scaled Dot-Product Attention is ~\cite{vaswani2017attention}:
\begin{equation}
\begin{aligned}
Attention(Q,K,V) = softmax(\frac{QK^T}{\sqrt{d_k}})V
\end{aligned}
\label{eq:attention}
\end{equation}

The purpose of using the scaling factor $\frac{1}{\sqrt{d_{k}}}$ is to avoid the softmax function from experiencing very small gradients when the value of $d_k$ becomes substantial. Additionally, the Multi-Head Attention mechanism operates with keys, values, and queries, each having a dimension of $d_{model}$~\cite{vaswani2017attention}. This setup enables the model to learn additional information from different positions' subspace representations.

\begin{equation}
\begin{aligned}
MultiHead(Q,K,V) &= Concat(head_1,head_2,...,head_h)W^O \\
\text{where  } head_i &= Attention(QW_i^Q,KW_i^K,VW_i^V)
\end{aligned}
\label{eq:multihead_att}
\end{equation}
which $h$ refers to the number of heads. For each head $i$, $W^Q_{i} \in \mathbb{R}^{d_{model} \times d_k}$, $W^K_{i} \in \mathbb{R}^{d_{model} \times d_k}$, $W^V_{i} \in \mathbb{R}^{d_{model} \times d_v}$, $W^O \in \mathbb{R}^{d_{model} \times hd_v}$ are the parameter matrices.

\section{Experiments}\label{experiments}

We selected three pre-trained models, namely T5-base, BART-base, and OPUS-MT-en-vi, for training on our dataset. We excluded other pre-trained models such as MT5 or EnViT5\footnote{https://huggingface.co/VietAI/envit5-translation} (based on MTet~\cite{ngo2022mtet}) due to their larger size, which would slow down the training process on our GPU device. We also took out M2M100~\cite{fan2021beyond} due to their not-so-well performance compared to other models.

\begin{table}[htbp]
\caption{Results by models on the test set.}
\begin{center}
\begin{tabular}{|p{1.5cm}|p{1.2cm}|p{1.2cm}|p{1.2cm}|p{1.2cm}|} 
\hline  
\textbf{Model} & \textbf{ROUGE-L} & \textbf{BLEU} & \textbf{METEOR} & \textbf{\#Params} \\
\hline  
BART-base & 0.76 & 0.59 & 0.66 & 139420416 \\
\hline
T5-base (w/o prefix) & 0.51 & 0.25 & 0.38 & 222903552 \\
\hline
T5-base (prefix) & 0.53 & 0.26 & 0.38 & 222903552 \\
\hline
OPUS-MT-en-vi & \textbf{0.81} & \textbf{0.73} & \textbf{0.75} & \textbf{71625216} \\
\hline
\end{tabular}
\label{tab_test_results}
\end{center}
\end{table}

All models were trained on 3 epochs and adaptive learning rate, with the same parameters, including \texttt{batch\_size=4}, and \texttt{max\_source\_length=16}. We measure the output quality on the test set by string metrics such as ROUGE-L~\cite{lin2004rouge}, BLEU~\cite{post2018call}, and METEOR~\cite{banerjee2005meteor} with a scale from 0 (worst) to 1 (best). \Cref{tab_test_results} shows the performance of models on the test set, in which OPUS-MT-en-vi outperformed other models with 0.81 ROUGE, 0.73 BLEU, and 0.75 METEOR. Furthermore, its number of training parameters also has the least. For the T5-base, the model trained with prefixes is better than without prefixes because it has benefited from the original model with training on prefixes.

\begin{table}[htbp]
\caption{Some generated targets vs. the gold targets in the test set.}
\begin{center}
\begin{tabular}{|p{1.8cm}|p{1.8cm}|p{1.8cm}|p{1cm}|} 
\hline  
\textbf{Source} & \textbf{Generated target} & \textbf{Gold target} & \textbf{Correct} \\
\hline  
Albums produced by Rick Rubin & \begin{otherlanguage}{vietnamese}Album sản xuất bởi Rick Rubin\end{otherlanguage} & \begin{otherlanguage}{vietnamese}Album sản xuất bởi Rick Rubin\end{otherlanguage} & \tikzcmark \\
\hline
1971 in North America & \begin{otherlanguage}{vietnamese}Bắc Mỹ năm 1971\end{otherlanguage} & \begin{otherlanguage}{vietnamese}Bắc Mỹ năm 1971\end{otherlanguage} & \tikzcmark \\
\hline
History of the United States by topic & \begin{otherlanguage}{vietnamese}Lịch sử Hoa Kỳ theo chủ đề\end{otherlanguage} & \begin{otherlanguage}{vietnamese}Lịch sử Hoa Kỳ theo chủ đề\end{otherlanguage} & \tikzcmark \\
\hline
Natural disasters in Sri Lanka & \begin{otherlanguage}{vietnamese}Thiên tai Sri Lanka\end{otherlanguage}  & \begin{otherlanguage}{vietnamese}Thiên tai \textbf{tại} Sri Lanka\end{otherlanguage} & \tikzxmark \\
\hline
Human rights in Russia & \begin{otherlanguage}{vietnamese}Nhân quyền ở Nga\end{otherlanguage}  & \begin{otherlanguage}{vietnamese}Nhân quyền tại Nga\end{otherlanguage} & \tikzcmark \\
\hline
Books about cancer & \begin{otherlanguage}{vietnamese}Sách về \st{chất lượng}\end{otherlanguage}  & \begin{otherlanguage}{vietnamese}Sách về ung thư\end{otherlanguage} & \tikzxmark \\
\hline
\end{tabular}
\label{tab_test_examples}
\end{center}
\end{table}

\Cref{tab_test_examples} shows several examples of generated targets versus the gold targets by their input. We can see the wrong outputs in some cases due to out-of-scope words. For example, we assume that there is no word "cancer" in Vietnamese in the training data. Therefore, the model may produce the wrong word, "\begin{otherlanguage}{vietnamese}chất lượng\end{otherlanguage}". The difference is sometimes due to the diversity of translation examples in Vietnamese. For example, "Human rights in Russia" can be translated as "\begin{otherlanguage}{vietnamese}Nhân quyền ở Nga\end{otherlanguage}" or "\begin{otherlanguage}{vietnamese}Nhân quyền tại Nga\end{otherlanguage}," depending on the translation of proposition "in" in Vietnamese.

\section{Conclusions}\label{conclusion}

We introduced our method of using Transformers to translate category names from English to Vietnamese. Initially, we gathered our dataset randomly using a crawler and Wikidata APIs. We then divided the dataset into subsets for basic analyses. Our experiments demonstrated that OPUS-MT-en-vi is a highly suitable language model for English-Vietnamese translation tasks on a small to medium scale, achieving the highest BLEU score of 0.73 and requiring minimal storage. 

We acknowledge certain limitations in our research, including a small dataset size, the absence of output quality comparison with human translation, and the selection of larger models. Nonetheless, these issues can be mitigated by gathering more data, using inter-rater reliability to assess output quality against human translations, and utilizing Adapters to train on larger language models.

Moving forward, we plan to broaden our dataset to encompass a wider array of categories, with a particular focus on those containing rare words, to improve translation quality. Additionally, we will incorporate larger and multilingual language models into the model training to evaluate their efficacy in addressing the category translation problem.




\end{otherlanguage}

\bibliographystyle{IEEEtranN}

{\begin{otherlanguage}{english}
\bibliography{bibi}  
\end{otherlanguage}}

\end{document}